\documentclass[runningheads]{llncs}
 
\usepackage{eccv}

\usepackage{eccvabbrv}

\usepackage{booktabs}
\usepackage{chapterbib}
\usepackage{float}
\usepackage{multirow}
\usepackage{graphicx, caption}
\graphicspath {{./fig/}} 
\usepackage{sidecap} 
\usepackage{amsmath,amsfonts,amssymb}
\usepackage{bm}
\usepackage{textcomp}
\usepackage{stfloats}
\usepackage{verbatim}
\usepackage[normalem]{ulem}
\usepackage[section]{placeins}
\usepackage{enumerate}
\usepackage{enumitem}

\usepackage{algorithm}  
\usepackage{algpseudocode}

\usepackage[flushleft]{threeparttable}

\usepackage[accsupp]{axessibility}  

\usepackage[colorlinks]{hyperref}

\usepackage{orcidlink}

\begin{document}

\title{Enhanced Event-Based Video Reconstruction with Motion Compensation} 


\author{Siying Liu\inst{1}\orcidlink{0000-0003-3533-8768} \and
Pier Luigi Dragotti \inst{1}\orcidlink{0000-0002-6073-2807}}

\authorrunning{S.~Liu and P.L.~Dragotti}

\institute{Communication and Signal Processing Group, Department of Electrical and Electronic Engineering, Imperial College London, SW7 2AZ London, UK \email{siying.liu20@imperial.ac.uk}}

\maketitle

\begin{abstract}


Deep neural networks for event-based video reconstruction often suffer from a lack of interpretability and have high memory demands. A lightweight network called CISTA-LSTC has recently been introduced showing that high-quality reconstruction can be achieved through the systematic design of its architecture. However, its modelling assumption that input signals and output reconstructed frame share the same sparse representation neglects the displacement caused by motion. To address this, we propose warping the input intensity frames and sparse codes to enhance reconstruction quality. A CISTA-Flow network is constructed by integrating a flow network with CISTA-LSTC for motion compensation. The system relies solely on events, in which predicted flow aids in reconstruction and then reconstructed frames are used to facilitate flow estimation. We also introduce an iterative training framework for this combined system. Results demonstrate that our approach achieves state-of-the-art reconstruction accuracy and simultaneously provides reliable dense flow estimation. Furthermore, our model exhibits flexibility in that it can integrate different flow networks, suggesting its potential for further performance enhancement. 

\end{abstract}

\section{Introduction}
\label{sec:introduction}

Event cameras, closely resembling human eyes and offering enhanced energy efficiency, have drawn significant interest in the field of computer vision \cite{gallego_event-based_2020, tayarani-najaran_event-based_2021}. These cameras sense changes in brightness and asynchronously output pixel-wise event streams. This working principle enables the generation of event data characterized by low latency, high temporal resolution and high dynamic range. Due to the unique way in which information is encoded, events-to-video (E2V) reconstruction becomes a fundamental research topic.

Many studies have explored various approaches to reconstruct intensity videos with high frame rates, high dynamic range, and reduced motion blur. Traditional model-based methods \cite{brandli_real-time_2014, bardow_simultaneous_2016, reinbacher_real-time_2016, scheerlinck_continuous-time_2018} leverage the relationship between events and intensity as priors in their models, but their performance is limited as these priors are not rich enough to diverse scenarios. In contrast, learning-based methods have achieved high-quality video reconstruction. These methods usually employ Unet-like architectures with recurrent or specially designed blocks to enhance temporal consistency between frames, such as E2VID \cite{rebecq_events--video_2019, rebecq_high_2021}, FireNet \cite{scheerlinck_fast_2020}, SPADE-E2VID \cite{cadena_spade-e2vid_2021}, and HyperE2VID \cite{ercan_hypere2vid_2023}. Generative adversarial networks \cite{wang_event-based_2019, su_event-based_2020} and Transformer-based networks like ET-Net \cite{weng_event-based_2021} have also been applied for E2V reconstruction. However, many of these networks require large memory consumption and the underlying principles behind them remain unclear. Zhu et al. \cite{zhu_event-based_2022} transformed the UNet-like model into a spiking neural network to reduce computational costs, but it still relies on deep layers to maintain high performance. Recently, Liu \etal~\cite{liu_convolutional_2022, liu_sensing_2023} introduced a lightweight convolutional ISTA network with long short-term temporal consistency (CISTA-LSTC), which is systematically designed based on the notion of sparse representation. This systematic design approach provides us insights on how performance can be further enhanced. In this work, we revisit its modelling assumption between events and intensity frames. We find that the assumption of shared common sparse codes between input and output signals neglects the displacement caused by motion. We therefore propose a modification based on incorporating optical flow for motion compensation in the inputs of CISTA-LSTC.


Some studies have leveraged the relationship between motion compensated events and the gradients of intensity frames to estimate optical flow and retrieve intensity simultaneously. For example, Paredes-Valles \etal~\cite{paredes-valles_back_2021} proposed a self-supervised framework based on photometric constancy between the warped events and the gradients of the reconstructed image. Similarly, Zhang et al. \cite{zhang_formulating_2022} used estimated flow to warp events initially and then employed learning-based regularizers to solve the linear problem for image reconstruction. For flow estimation from events, the contrast maximization framework \cite{gallego_unifying_2018, stoffregen_event_2019, avidan_secrets_2022} can estimate motion from event frames, but careful selection of objective function is required. Deep networks such as EV-FlowNet \cite{zhu_ev-flownet_2018} and its variations \cite{vedaldi_spike-flownet_2020, ding_spatio-temporal_2022, ponghiran_event-based_2023} have shown advancements in flow estimation for long sequences. However, these networks can only estimate sparse flow in areas where events occur. Recent studies have explored RAFT-like architectures \cite{teed_raft_2020} to achieve high-quality dense flow estimation. ERAFT \cite{gehrig_e-raft_2021} captures correlations between two consecutive groups of events, while DCEIFlow \cite{wan_learning_2022} utilizes the fusion of events and corresponding intensity frame to facilitate continuous and dense flow estimation. 


In this work, we introduce a CISTA-Flow network to enhance reconstruction with motion compensation by integrating CISTA-LSTC with a flow estimation network, particularly utilizing the DCEIFlow \cite{wan_learning_2022}. In this model, the reconstructed image and corresponding events are fed into DCEIFlow to estimate flow. Subsequently, the flow is utilized to warp both the previously reconstructed frame and sparse codes as the inputs of CISTA-LSTC for current reconstruction. We also establish an iterative training framework for this hybrid system. Our CISTA-Flow network achieves state-of-the-art reconstruction quality on both simulated and real event datasets, while enabling dense flow estimation simultaneously. Furthermore, this architecture offers flexibility to integrate other event-based flow networks, enabling easy adaptation of the network. Overall, the main contributions of this study are as follows:
\begin{enumerate}
    \item Improve the CISTA-LSTC network by integrating an optical flow estimation network to address motion compensation, leading to high reconstruction quality.
    \item Achieve simultaneous optical flow estimation and video reconstruction solely from events, without the requirement for additional priors. 
    \item Systematically design the CISTA-Flow based on sparse representation, resulting in an interpretable system with a flexible adaptation architecture.
\end{enumerate}

\section{Methodology}
\label{sec:method}
Consider an event stream $\{e_i, i=1,2,\cdots, N\}$, occurring over a duration of $T$ seconds. Each event $e_i$ can be mathematically represented as a function $s(x,y,t)=p_i\delta (x-x_i, y-y_i, t-t_{i})$. Here, $\delta(\cdot)$ represents the Dirac function, $(x_i, y_i)^\mathrm{T}$ denotes the spatial coordinates of the event, $t_i$ is its timestamp, and $p_i=\pm 1$ indicates the polarity of its brightness change. In E2V reconstruction, aggregating a group of events is essential to gather sufficient information for each reconstruction. Typically, a 3D event voxel grid $\bm E\in \mathbb{R}^{B\times H\times W}$ \cite{zhu_ev-flownet_2018} is created by accumulating event polarities into $B$ temporal bins, where $H$ and $W$ represent the height and width of the image, respectively. Each polarity is distributed linearly into the two closest temporal voxels. 

In this study, our objective is to reconstruct the frame $\hat{\bm I}_t \in \mathbb{R}^{H\times W}$ at time instance $t$ by utilizing the previously reconstructed frame $\hat{\bm I}_{t-1}\in \mathbb{R}^{H\times W}$ and the corresponding event voxel grid $\bm E_{t-1}^{t}\in \mathbb{R}^{B\times H\times W}$. In \cref{sec:prob-cista}, we revisit CISTA-LSTC network for E2V reconstruction and analyse its limitations. We then introduce a CISTA-Flow network in \cref{sec:rec-flownet}, which integrates the CISTA-LSTC network with a flow estimation network to address motion compensation for video reconstruction. Finally, \cref{sec:train-framework} presents an iterative training framework for the hybrid system.

\vspace{-0.5em}
\subsection{Overview of CISTA-LSTC Network}\label{sec:prob-cista}


\subsubsection{Overview of CISTA-LSTC}
\begin{figure}[tb]
    \centering
    \includegraphics[width=1.0\textwidth]{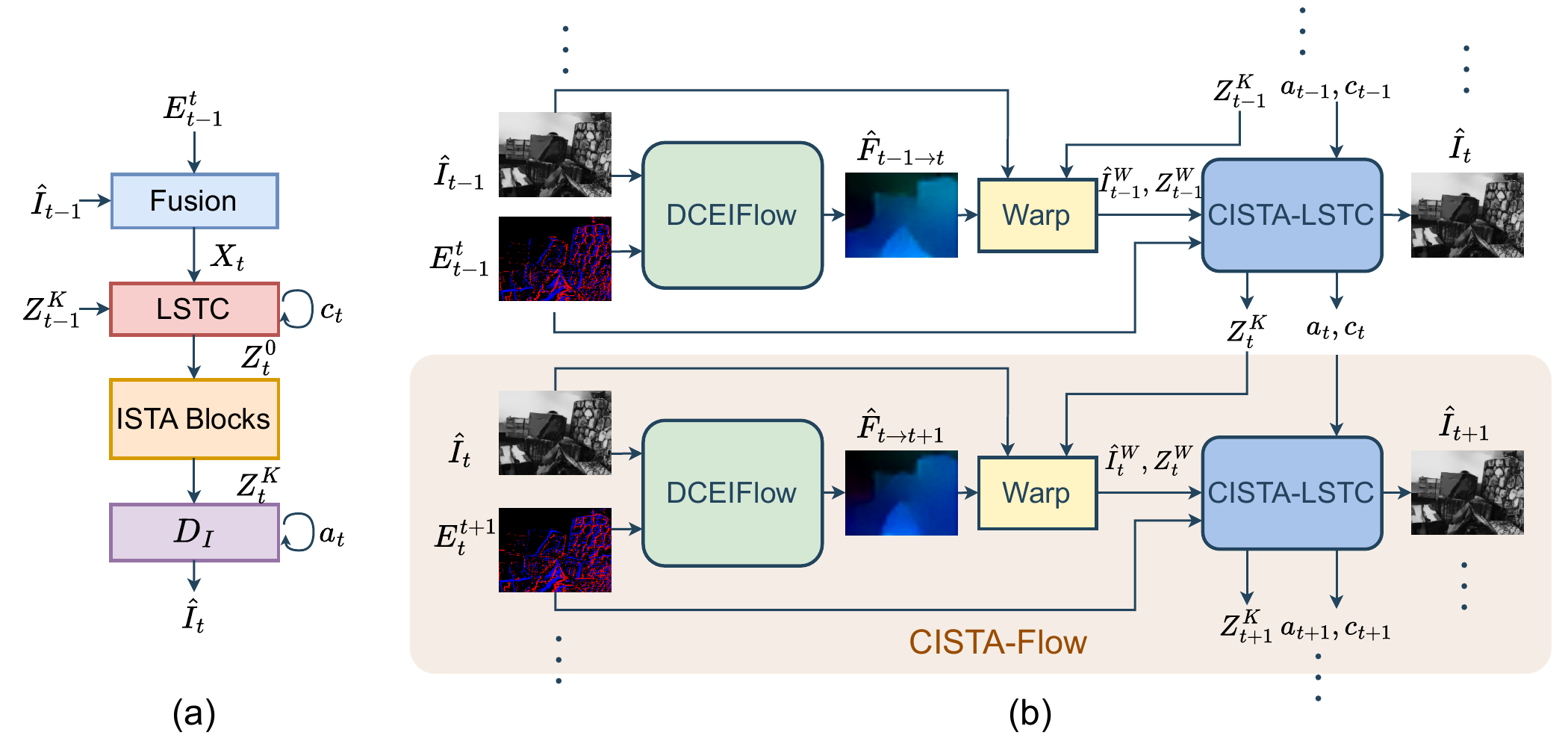}
    \caption{Recursive CISTA-Flow architecture. (a) Original CISTA-LSTC network. (b) CISTA-Flow network. The DCEIFlow network leverages the previously reconstructed frame $\hat{\bm I}_{t-1}$ and event voxel grid ${\bm E}_{t-1}^t$ to estimate the forward flow $\hat{\bm F}_{t-1\rightarrow t}$. This estimated flow is then utilized to warp $\hat{\bm I}_{t-1}$ and the sparse codes ${\bm Z}_{t-1}^K$ obtained from the previous reconstruction. Finally, CISTA-LSTC employs these warped inputs to reconstruct the frame $\hat{\bm I}_{t}$. The initial $\hat{\bm I}_{0}$ is set to 0.}
    \label{fig:cista-flow-arch}
    \vspace{-1em}
\end{figure}

The CISTA-LSTC network, as shown in \cref{fig:cista-flow-arch}(a), is designed based on sparse representation paradigm \cite{elad_sparse_2010,aharon_k-svd_2006}. Input events $\bm E_{t-1}^t$ and intensity frame $\hat{\bm I}_{t-1}$ are fused as a stack of feature maps $\bm X_t$. It is assumed that the input $\bm X_t$ shares the same sparse representation with the output frame $\hat{\bm I}_t$, i.e. $\bm X_t = \bm D_X \bm Z^*$, $\hat{\bm I}_t = \bm D_I \bm Z^*$, where $\bm D_X, \bm D_I$ are corresponding dictionaries and $Z^*$ represents sparse feature maps which are in common. The sparse coding problem can be solved by the iterative shrinkage thresholding algorithm (ISTA) \cite{daubechies_iterative_2004}. The CISTA network is then developed using algorithm unfolding strategy \cite{gregor_learning_2010, monga_algorithm_2021} by unfolding the iteration in the ISTA. In the CISTA-LSTC network, the input $\bm X_t$ is processed by a long short-term temporal consistency (LSTC) unit to initialise sparse codes $\bm Z_t^0$. Following that, the $K$ unfolded ISTA blocks are employed to optimise sparse codes $\bm Z_t^k$ at each iteration. Finally, a long short-term reconstruction consistency (LSRC) block serves as the synthesis dictionary $\bm D_I$ to reconstruct the intensity frame $\hat{\bm I}_{t}$. Both the LSTC and LSRC are recurrent blocks, aimed at enhancing temporal consistency in reconstruction, updating states $\bm c_{t}$ and $\bm a_t$ respectively. In particular, the LSTC block takes the sparse code estimated at the previous reconstruction $\bm Z_{t-1}^K$ and current input $\bm X_t$ to initialise sparse code $\bm Z_t^0$, preserving the temporal consistency in sparse codes $\bm Z$.


\vspace{-1em} 
\subsubsection{Motion compensation for input frame and sparse codes}
The assumption of shared sparse codes between the input $\bm X_t$ and the output $\hat{\bm I}_{t}$ holds because their structures are similar. However, there exists a motion displacement between the previously reconstructed image $\hat{\bm I}_{t-1}$ and the output image $\hat{\bm I}_{t}$. Furthermore, the similarity between the input $\bm X_t$ and sparse codes $\bm Z_{t-1}^K$ are utilized to preserve temporal consistency in sparse codes. This consistency is also affected by the spatial displacement introduced by motion. Although the additional information from events is expected to compensate for the displacement, we hypothesize that compensating for the motion from instant $t-1$ to $t$ could enhance performance. If the motion is known, warping input frame $\hat{\bm I}_{t-1}$ and the previous sparse codes $\bm Z_{t-1}^K$ to serve as the inputs may lead to improved reconstruction accuracy. Specifically, the inputs $\hat{\bm I}_{t-1}$ and $\bm Z_{t-1}^K$ are warped using the known optical flow $\bm F_{t-1\rightarrow t}$, as described by the following equations:
\begin{equation}
    \begin{aligned}
        \hat{\bm I}_{t-1}^W &= \mathcal{W}_f(\hat{\bm I}_{t-1}, \bm F_{t-1\rightarrow t}), \\
        {\bm Z}_{t-1}^{W} &= \mathcal{W}_f({\bm Z}_{t-1}^{K}, \bm F^d_{t-1\rightarrow t}),
    \end{aligned}
    \label{eq:warp_input}
\end{equation}
where the $\mathcal{W}_f(\bm I, \bm F)$ denotes the forward warping function to warp frame $\bm I$ using flow $\bm F \in \mathbb{R}^{2\times H\times W}$. Each element in the flow map represents vertical and horizontal displacement in pixel. Note that the spatial size of ${\bm Z}_{t-1}^{K}$ is $\frac{H}{2} \times \frac{W}{2}$, thus we employ the bilinearly downsampled flow map, denoted as $\bm F^d_{t-1\rightarrow t}$. 

\begin{figure}[tb]
    \centering
    \subfloat[No warp]{
    \begin{minipage}[c][\height][c]{0.18\linewidth}
    \centering
     \includegraphics[width=\linewidth]{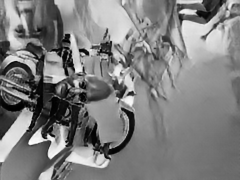}
    \includegraphics[width=\linewidth]{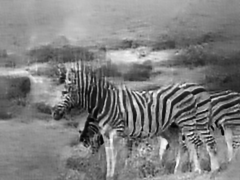}
    \end{minipage}
    \label{subfig:no-warp}
    }
    \subfloat[Warp I]{
    \begin{minipage}[c][\height][c]{0.18\linewidth}
    \centering
    \includegraphics[width=\linewidth]{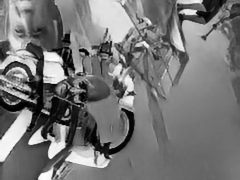}
    \includegraphics[width=\linewidth]{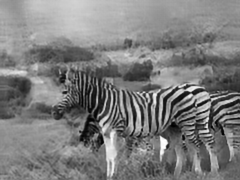}
    \end{minipage}
    \label{subfig:warp-I}
    }
    \subfloat[Warp I+Z]{
    \begin{minipage}[c][\height][c]{0.18\linewidth}
    \centering
    \includegraphics[width=\linewidth]{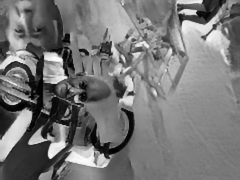}
    \includegraphics[width=\linewidth]{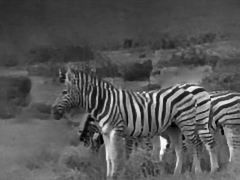}
    \end{minipage}
    \label{subfig:warp-I-Z}
    }
    \caption{Reconstruction results with different warped frames. "Warp I" denotes the model with warped $\hat{\bm I}_{t-1}$ and "Warp I+Z" denotes the model with warped $\hat{\bm I}_{t-1}$ and ${\bm Z}_{t-1}^{K}$.}
    \label{fig:compate-warp-gt}
    \vspace{-1em}
\end{figure}

To validate our hypothesis, we conducted an experiment to reconstruct images using different warped inputs. Experimental details are described in \cref{sec:results-diff-warped-inputs}. As shown in \cref{fig:compate-warp-gt}, models utilizing warped $\bm I$ and $\bm Z$ yield enhanced details and sharper edges compared to the model without warping. Therefore, we consider learning to estimate optical flow from events, serving as a motion compensation module to enhance the quality of video reconstruction.


\subsection{Events-to-Video Reconstruction with Motion Compensation}\label{sec:rec-flownet}

In this section, we introduce a CISTA-Flow network designed to enhance E2V reconstruction with motion compensation. This network integrates both a flow estimation network and a CISTA-LSTC network for E2V reconstruction. The overall recursive architecture is shown in \cref{fig:cista-flow-arch}(b). The flow network takes the event voxel grid $\bm E_{t-1}^{t}$ and the reconstructed frame $\hat{\bm I}_{t-1}$ as inputs to estimate the forward optical flow $\hat{\bm F}_{t-1\rightarrow t}$ from instant $t-1$ to $t$. Subsequently, this estimated flow is utilized to warp both $\hat{\bm I}_{t-1}$ and ${\bm Z}_{t-1}^K$ as the inputs for the CISTA-LSTC network. Finally, the reconstructed frame $\hat{\bm I}_t$ is obtained. The initial input frame $\hat{\bm I}_0$ is initialised with $\bm 0$. Through the iterative prediction, the CISTA-Flow network demonstrates the ability to accurately predict flow and reconstruct high-quality intensity frames using events only. The two-stage estimation process is as follows.


\vspace{-0.5em}
\subsubsection{Flow estimation using reconstructed images and events}\label{sec:flownet}

Various architectures of the flow network can be employed, and we typically utilize the DCEIFlow network \cite{wan_learning_2022} to estimate reliable dense flow. The DCEIFlow network, based on a RAFT architecture \cite{teed_raft_2020} initially developed for two-frame-based flow estimation, utilizes a single intensity frame and event streams as inputs. It employs the fused feature embedding as pseudo feature maps for $\bm I_2$, along with the feature maps for $\bm I_1$, to generate continuous and dense flow via their correlation volume and an iterative updater. In our system, ground truth (GT) intensity image is not available but the CISTA-LSTC network can provide reconstructed frame as the complimentary information. We therefore fine-tune the DCEIFlow to predict flow based on reconstructed images and corresponding events. 
The process of estimating flow $\hat{\bm F}_{t-1\rightarrow t}$ can be expressed as follows:
\begin{equation}
    \hat{\bm F}_{t-1\rightarrow t} = \text{DCEIFlow}(\bm E_{t-1}^{t}, \hat{\bm I}_{t-1}).
    \label{eq:flow-est}
\end{equation}

\vspace{-0.5em}
\subsubsection{Video reconstruction using warped frames and sparse codes}\label{sec:recnet}

With the predicted flow $\hat{\bm F}_{t-1\rightarrow t}$, we can warp the reconstructed image $\hat{\bm I}_{t-1}$ and sparse codes $\bm Z_{t-1}^K$ to serve as inputs along with ${\bm E}_{t-1}^t$ for the subsequent reconstruction. This process can be expressed as follows:
\begin{equation}
    \begin{aligned}
        \hat{\bm I}_{t-1}^W &= \mathcal{W}_f(\hat{\bm I}_{t-1}, \hat{\bm F}_{t-1\rightarrow t}), \\
        {\bm Z}_{t-1}^{W} &= \mathcal{W}_f({\bm Z}_{t-1}^{K}, \hat{\bm F}^d_{t-1\rightarrow t}), \\
        [\hat{\bm I}_{t}, {\bm Z}_{t}^K, \bm a_{t}, \bm c_{t}] &= \text{CISTA-LSTC}({\bm E}_{t-1}^t, \hat{\bm I}_{t-1}^W, {\bm Z}_{t-1}^{W}, \bm a_{t-1}, \bm c_{t-1}),
    \end{aligned}
    \label{eq:rec_warp}
\end{equation}

\subsection{Iterative Training Framework}\label{sec:train-framework}

Simultaneously training flow and reconstruction networks poses challenges as their accuracy is interdependent. The quality of flow affects reconstruction and vice versa. To address this problem, we propose an iterative training framework based on pretrained CISTA-LSTC and DCEIFlow networks. 

\begin{figure}[tb]
    \centering
    \includegraphics[width=\linewidth]{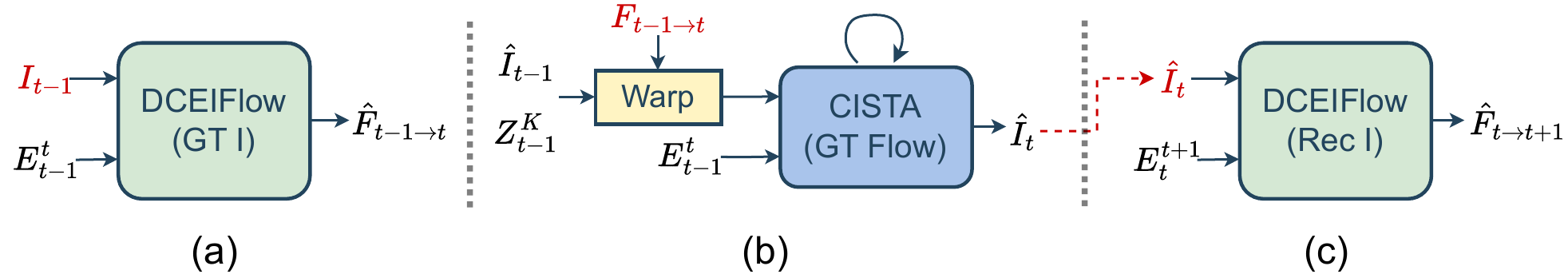}
    \caption{Separate training for (a) DCEIFlow using ground truth input frame and (b) CISTA-LSTC using ground truth flow, and additional training for (c) DCEIFlow (Rec I) using reconstructed frames generated by CISTA (GT Flow). }
    \label{fig:it-train}
    \vspace{-1em}
\end{figure}

The training process can be divided into three stages. First, as shown in \cref{fig:it-train}(a)(b), DCEIFlow and CISTA-LSTC undergo independent training for 50 and 60 epochs respectively. Specifically, the original DCEIFlow with the GT intensity frame as input is trained based on the pretrained model provided by authors using our dataset, denoted as DCEIFlow (GT I). The CISTA-LSTC network is trained from scratch using warped inputs generated by GT flow, referred to as CISTA (GT Flow). 

Subsequently, we conduct additional training for 20 epochs on the DCEIFlow to make it adaptable to the reconstructed frames, denoted as DCEIFlow (Rec I), as illustrated in \cref{fig:it-train}(c). At this stage, the reconstructed frame is obtained from the pretrained CISTA (GT Flow) model.

Finally, after separate training of the reconstruction and flow networks, we iteratively fine-tune both networks to obtain the CISTA-Flow network depicted in \cref{fig:cista-flow-arch}(b). This process allows each network to adapt to the predictions from the other. During training, the CISTA and DCEIFlow networks are fixed alternately. Specifically, we fix the parameters of CISTA to generate reconstructed frames when training DCEIFlow, and vice versa. The iterative training involves updating one of the two networks every 2 epochs over 20 epochs.

During training, the loss is computed after each reconstruction or flow estimation, but weights are updated at the end of the sequence. Each sequence contains $L$ frames. Below, we introduce the loss functions employed for reconstruction and flow estimation. Note that a weighing term $\bm M_{t}=\exp(-\alpha_M |\mathcal{W}_f({\bm I}_{t-1}, {\bm F}_{t-1\rightarrow t})-\bm I_t|^2)$ utilized in certain loss functions, where $\alpha_M=50$. This term is used to mitigate the impact of significant errors in the ground truth flow and warped frames \cite{rebecq_high_2021}.

\vspace{-0.5em}
\subsubsection{Reconstruction loss}
The loss function for the $t$th reconstructed frame is a combination of $\ell_1$, structural similarity (SSIM) and perceptual loss (LPIPS) \cite{zhang_unreasonable_2018}, as follows:
\begin{equation}
    L_{rec}^t =   \|\bm \hat{\bm I}_{t}-{\bm I}_t \|_1 + (1-\text{SSIM}(\hat{\bm I}_{t},{\bm I}_t)) + \text{LPIPS}(\hat{\bm I}_{t},{\bm I}_t),
\end{equation}
where $\|\cdot\|_1$ represent the $\ell_1$ norm and the VGG \cite{simonyan_very_2015} is used in the LPIPS loss. 

In addition, a temporal consistency loss \cite{rebecq_high_2021} is introduced to preserve the temporal consistency of intensity using flow, which is given by:
\begin{equation}
    L_{tc}^t = \|\bm M_{t} \left (\mathcal{W}_f(\hat{\bm I}_{t-1}, {\bm F}_{t-1\rightarrow t})-\hat{\bm I}_t \right)  \|_1
\end{equation}

At the end of each sequence, the reconstruction loss $L_R$, which aggregates $L_{rec}^t$ and $L_{tc}$ across the entire sequence, is given by:
\begin{equation}
    L_{R} = \sum_{t=1}^L L^{t}_{rec} + \lambda_{tc}\sum_{t=L_0}^L L^t_{tc},
\end{equation}
where $\lambda_{tc}=5$, and $L_0=3$ to skip the first few frames to allow convergence.

\vspace{-0.5em}
\subsubsection{Flow loss}
For the flow estimation, DCEIFlow has $R$ iterations. We calculate the loss for the flow map at each iteration, utilizing a weighted sum of the $L_1$ loss between the estimated flow and the GT flow map, as follows:
\begin{equation}
    L^t_f = \sum_{i=1}^R w_i \|\bm M_{t}^i (\hat{\bm F}^i_{t-1\rightarrow t}-{\bm F}^i_{t-1\rightarrow t} )\|_1,
\end{equation} 
where ${\bm F}^i_{t-1\rightarrow t}$ represents the bilinear downsampled flow map for the $i$th iteration, and $\bm M_{t}^i$ is downsampled accordingly. Moreover, the weight for the $i$th iteration is $w_i=\phi^{R - i - 1}$, where $\phi=0.8$ \cite{wan_learning_2022}.

Furthermore, a photometric loss for warped GT frames is included:
\begin{equation}
    L^t_{photo} = \sum_{i=1}^R w_i \|\mathcal{W}_f({\bm I}^i_{t-1}, \hat{\bm F}^i_{t-1\rightarrow t})-\bm I^i_t\|_1.
\end{equation}

Finally, the flow loss $L_F$ is calculated over the whole sequence and we set $\lambda_p=1$:
\begin{equation}
    L_{F} = \sum_{t=1}^L (L_f^t+ \lambda_p L^t_{photo}).
\end{equation}


\section{Numerical Results}
\label{sec:results}

\subsection{Experimental Settings and Training Details}
\label{sec:exp-train}

\subsubsection{Datasets}
For the training dataset, we utilize the v2e simulator \cite{hu_v2e_2021} to generate events and corresponding flow from intensity frames. Using images from MS-COCO datasets \cite{lin_microsoft_2014}, we simulate videos with various motions and 2D affine transform at adaptive frame rate. It is important to note that the GT flow is acquired using the FlowNet from the pretrained Super-Slomo model in v2e, which estimates flow based on two intensity frames. Although exact motion parameters are known during data generation, the real flow introduces more occlusions in warped frames. The estimated flow can reduce intensity error between the warped frame $\bm I^W_t$ and the target frame $\bm I_t$. Therefore, we employ the estimated flow as GT flow during training, acknowledging that potential inaccuracies in flow estimation. The image size is set to $240 \times 180$. The dataset contains 1538 training sequences and 180 testing sequences. Each sequence has length $L=15$, with approximately $N_E=15,000$ events between consecutive frames. For further details, please refer to the supplementary material.

In addition to the simulated dataset, we evaluate reconstruction performance using real datasets, including the Event Camera Dataset (ECD) \cite{mueggler_event-camera_2017} and the High-Quality Frames (HQF) \cite{storegen_reducing_2020}. These were captured by DAVIS240C cameras ($240\times 180$). The ECD comprises 7 sequences, while the HQF contains 14 sequences with a wider range of motions and scene types. HQF generally presents higher-quality intensity images with reduced motion blur compared to ECD.

We also evaluated our model on challenging datasets such as the MVSEC \cite{zhu_multi_2018} and the HS-ERGB \cite{tulyakov_time_2021} datasets. The MVSEC dataset contains stereo event data captured from various vehicles, using a pair of DAVIS 346B cameras ($346\times 260$). We utilized data from the left camera only. The sequences include indoor and outdoor scenarios, as well as low-light conditions like night driving. The HS-ERGB dataset comprises sequences with high resolution and rapid motion, captured by the Prophesee Gen4 ($1280 \times 720$). Due to the absence of reliable HDR reference images in these datasets, we only present qualitative results.



\vspace{-0.5em}
\subsubsection{Training details}

The Adam optimizer and a batch size of 2 are used during training. For separate training of CISTA (GT Flow) and DCEIFlow (GT I), and also other reconstruction networks, the learning rate starts at 0.0001, decaying by 10\% every 10 epochs. For DCEIFlow (Rec I), a constant learning rate of 0.0001 is applied. During the iterative training of the two components, a learning rate of $3e^{-5}$ is maintained.

\vspace{-0.5em}
\subsubsection{Evaluation}
We assess our model using the EVREAL \cite{ercan_evreal_2023} evaluation framework for E2V reconstruction, which incorporates popular datasets and mainstream networks, allowing for consistent evaluation across different methods. We adapt it to suit our settings. During inference, the constraint on the number of events $N_E$ remains consistent with the training phase. However, because GT intensity images are only available at a fixed frame rate, we either divide events between two consecutive frames into multiple groups or combine events across several frames into a single group for the reconstruction and flow estimation.

For reconstruction evaluation, we utilize mean square error (MSE), SSIM and LPIPS computed using AlexNet \cite{krizhevsky_imagenet_2012}. For the ECD dataset, considering the reference frames are relatively dark, we normalize the images into the range of $[0,1]$ to enable comparison with the reconstructed images. No post-processing is applied to the reconstructed frames. 


For optical flow estimation, we use the average endpoint error (EPE) to evaluate the error between predicted and GT flow vector for simulated data sequences. EPE is defined as the Euclidean distance between the GT flow vector and predicted flow vector. We also define the percentage of outliers when the EPE is greater than 3 pixels and 5\% of the magnitude of the flow vector. We only evaluate the flow estimation on our simulated datasets for reference. Please refer to the supplementary material for more results.


\subsection{Results of Video Reconstruction}

\subsubsection{Comparison against other reconstruction networks}

We compare the reconstruction performance of the CISTA-Flow network with other advanced reconstruction networks, including Unet-based networks such as E2VID \cite{rebecq_events--video_2019, rebecq_high_2021}, FireNet \cite{scheerlinck_fast_2020}, SPADE-E2VID \cite{cadena_spade-e2vid_2021}, and HyperE2VID \cite{ercan_hypere2vid_2023}, as well as the transformer-based ET-Net \cite{weng_event-based_2021} and the original CISTA-LSTC network \cite{liu_sensing_2023}. We also assess enhanced versions, E2VID+ and FireNet+, provided by \cite{storegen_reducing_2020}. SSL-E2VID \cite{paredes-valles_back_2021} is another version of E2VID. We retrain E2VID, FireNet, and SPADE-E2VID using our training data for the same epochs and also evaluate the pretrained models provided by the authors. 

In \cref{tab:results-compare}, the average comparison results on the ECD, HQF, and the simulated dataset (SIM) are presented. It is evident that CISTA-Flow performs the best or the second best on most metrics, particularly on SSIM. Comparing CISTA-Flow with the original CISTA-LSTC, we observe a significant improvement in accuracy, indicating that the reconstructed image structure is closer to the ground truth when utilizing predicted flow.
 
\begin{table}[tbp]
  \centering
  \caption{Comparison results against other reconstruction networks. SIM denotes synthetic testing datasets. Key: \textbf{Best}/ \underline{\textit{Second best}}}
  \setlength{\tabcolsep}{1mm}
  \scriptsize
  \begin{threeparttable}
\begin{tabular}{l|ccc|ccc|ccc}
\hline
\multirow{2}[4]{*}{} & \multicolumn{3}{c|}{ECD}       & \multicolumn{3}{c|}{HQF}       & \multicolumn{3}{c}{SIM} \\
\cline{2-10}         & MSE$\downarrow$ & SSIM$\uparrow$ & LPIPS$\downarrow$ & MSE$\downarrow$ & SSIM$\uparrow$ & LPIPS$\downarrow$ & MSE$\downarrow$ & SSIM$\uparrow$ & LPIPS$\downarrow$ \\
\hline
E2VID+*\cite{storegen_reducing_2020} & \underline{\textit{0.043}} & 0.535    & \underline{\textit{0.227}} & 0.035    & 0.527    & \textbf{0.246} & 0.040    & 0.471    & 0.440 \\
E2VID\cite{rebecq_events--video_2019, rebecq_high_2021} & 0.088    & 0.549    & 0.267    & 0.093    & 0.516    & 0.305    & 0.061    & 0.591    & 0.300 \\
FireNet+*\cite{storegen_reducing_2020} & 0.056    & 0.465    & 0.275    & \underline{\textit{0.034}} & 0.477    & 0.293    & 0.053    & 0.412    & 0.398 \\
FireNet\cite{scheerlinck_fast_2020} & 0.060    & 0.523    & 0.278    & 0.044    & 0.506    & 0.315    & 0.043    & 0.560    & 0.345 \\
SPADE-E2VID*\cite{cadena_spade-e2vid_2021} & 0.071    & 0.488    & 0.327    & 0.075    & 0.380    & 0.499    & 0.052    & 0.489    & 0.495 \\
SPADE-E2VID\cite{cadena_spade-e2vid_2021} & 0.083    & 0.558    & 0.249    & 0.084    & 0.525    & 0.289    & 0.035    & 0.646    & 0.257 \\
SSL-E2VID*\cite{paredes-valles_back_2021} & 0.069    & 0.432    & 0.397    & 0.067    & 0.449    & 0.389    & 0.088    & 0.392    & 0.540 \\
ETNet*\cite{weng_event-based_2021} & 0.053    & 0.511    & 0.237    & \textbf{0.033} & 0.524    & \underline{\textit{0.253}} & 0.051    & 0.462    & 0.398 \\
HyperE2VID*\cite{ercan_hypere2vid_2023} & 0.050    & 0.532    & \textbf{0.225} & \underline{\textit{0.034}} & 0.525    & 0.255    & 0.052    & 0.453    & 0.475 \\
CISTA-LSTC\cite{liu_sensing_2023} & \textbf{0.038} & \underline{\textit{0.585}} & 0.229    & 0.041    & \underline{\textit{0.563}} & 0.271    & \textbf{0.024} & \underline{\textit{0.650}} & \underline{\textit{0.252}} \\
\hline
CISTA-Flow & 0.047    & \textbf{0.586} & \textbf{0.225} & \underline{\textit{0.034}} & \textbf{0.590} & 0.257    & \underline{\textit{0.028}} & \textbf{0.654} & \textbf{0.244} \\
\hline
\end{tabular}%
\begin{tablenotes}
      \scriptsize
      \item The models marked with an asterisk (*) are pretrained models provided by the authors.
    \end{tablenotes}
\label{tab:results-compare}%
\end{threeparttable}
\end{table}%

\begin{figure}[tb]
     \centering
    \includegraphics[width=0.9\linewidth]{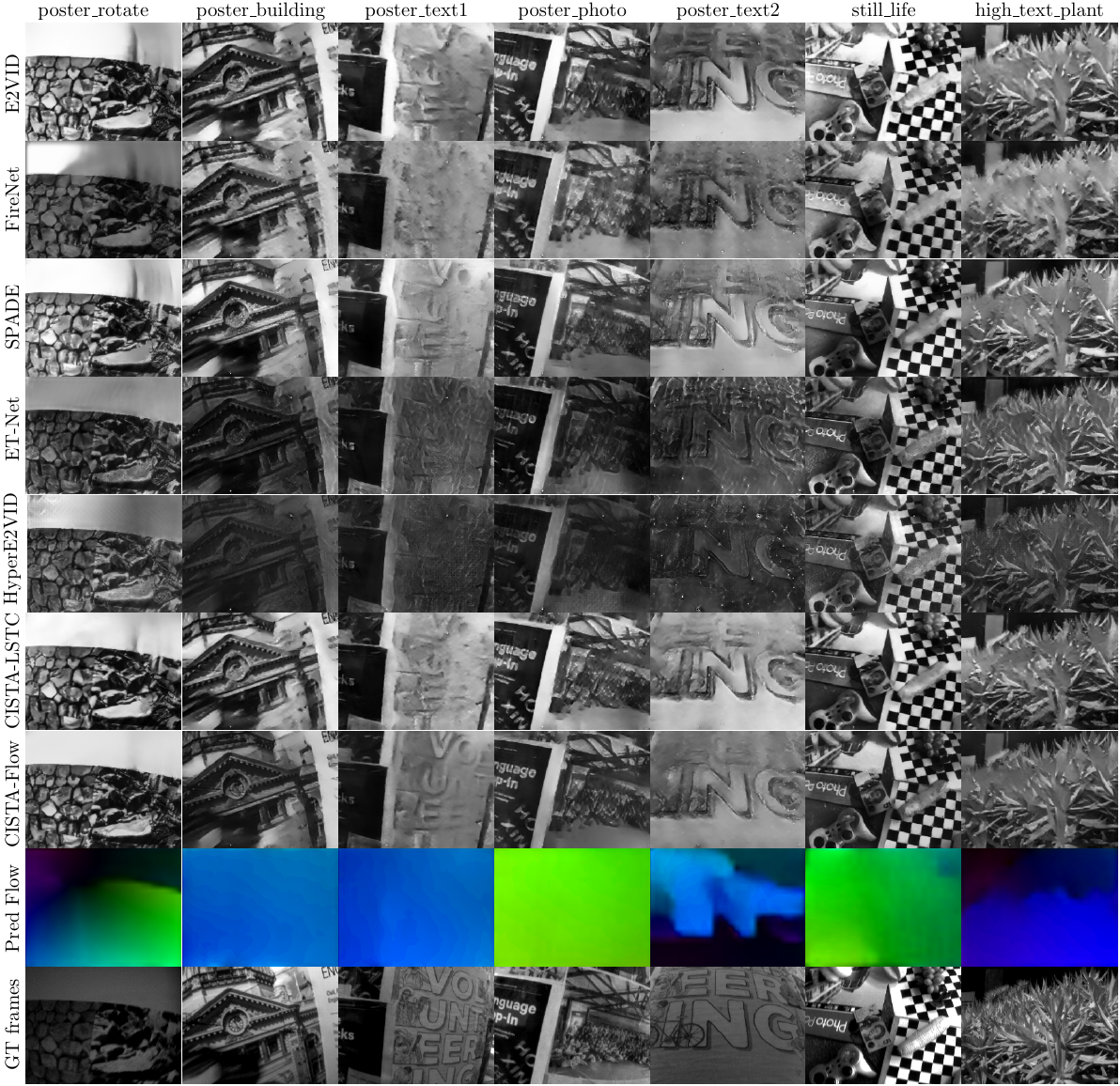}
     \caption{Comparison of video reconstruction between CISTA-Flow and other networks. } 
     \label{fig:compare}
\end{figure}

\cref{fig:compare} show some examples of reconstructed images from different networks and predicted flow from CISTA-Flow. The images produced by CISTA-Flow are sharper, with some details recovered thanks to motion compensation. ETNet and HyperE2VID effectively restores fine details and mitigates motion blur in many instances, yet the resulting reconstructed frames often appear underexposed and contain artifacts. For example, in scenes like poster\_text1 and poster\_text2, where texts are difficult to identify or ghosting occurs using other methods, our CISTA-Flow network enhances text visibility and reduces motion blur. Scenes like poster\_building, poster\_photo, still\_life, and high\_text\_plant highlight the recovery of fine details and sharper object edges. In the fast rotation scene of poster\_rotate, CISTA-Flow mitigates the bleeding effect that other networks exhibit. Finally, the predicted flow demonstrates the ability of CISTA-Flow to estimate reliable dense flow using events and reconstructed frames.

\vspace{-0.5em}
\subsubsection{Multi-objects scenes}
\begin{figure}[tb]
     \centering
    \includegraphics[width=0.8\linewidth]{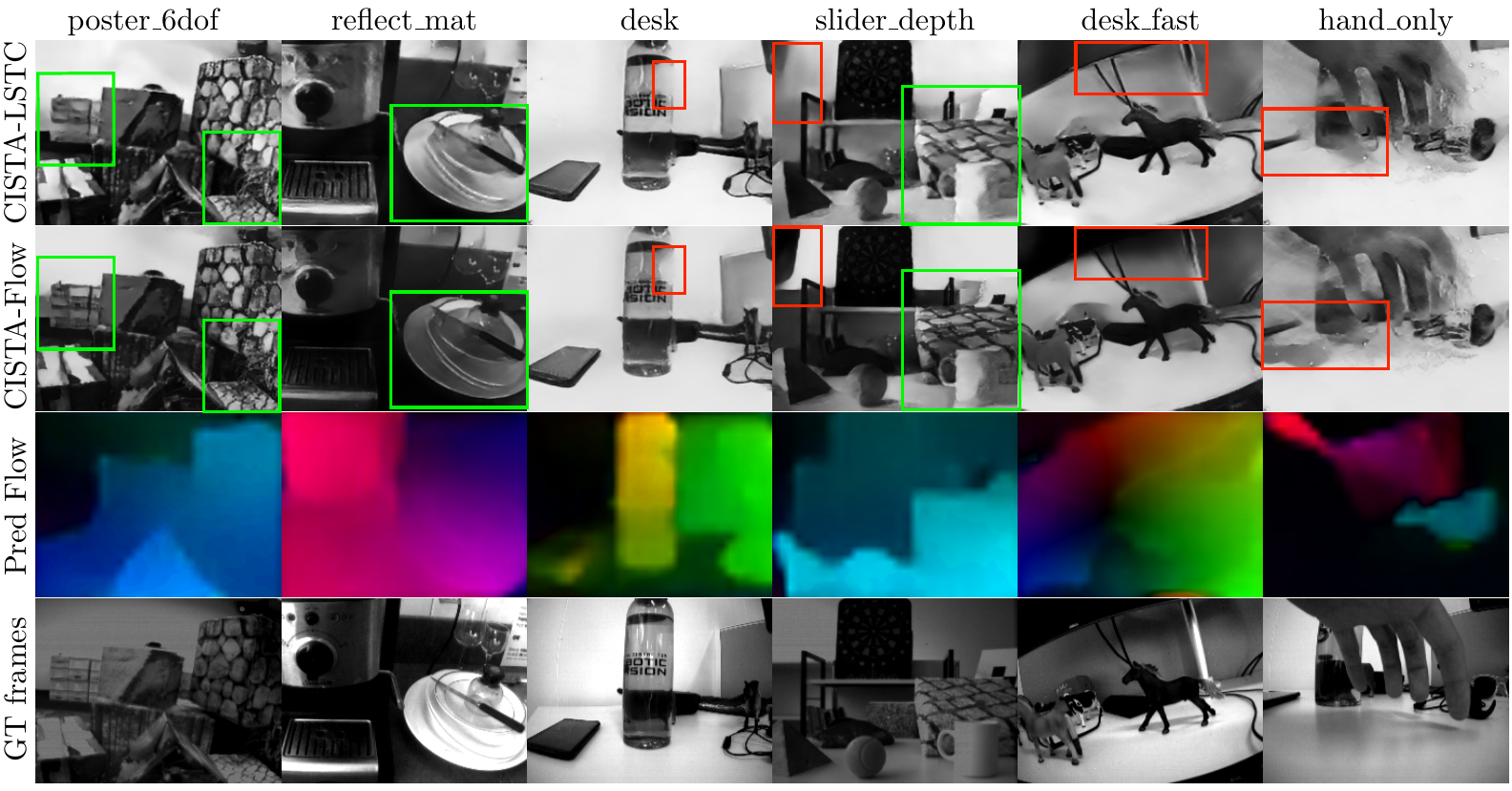}
     \caption{Reconstruction results of CISTA-LSTC and CISTA-Flow networks for scenes with multiple objects. The {\color{green} green} boxes highlight sharper edges and finer details improved by the estimated flow, whereas the {\color{red} red} boxes indicate areas with inaccuracies in reconstruction due to inaccurate flow.}
     \label{fig:compare-mul-obj}
\end{figure}

Apart from \cref{fig:compare}, we further present reconstruction results and related flow maps for scenes with multiple objects in \cref{fig:compare-mul-obj}. Different object positions result in diverse depths, leading to varying motion patterns. Flow maps demonstrate the capability of CISTA-Flow to reliably estimate dense flow in most scenarios. Comparing CISTA-Flow to the original CISTA-LSTC, accurate flow estimation leads to sharper edges and finer details in the resulting reconstructed frames. The green boxes highlight these enhancements.  

However, the accuracy of both reconstruction and flow estimation is interdependent. When reconstruction fails due to a lack of events or excessive noise, it affects the accuracy of optical flow, consequently impacting subsequent frames. In \cref{fig:compare-mul-obj}, areas with inaccuracies in CISTA-Flow reconstruction are highlighted by red boxes. In the desk scene, where the bottle is transparent and events are sparse around the edges, the estimated flow is inaccurate around this area and therefore affect reconstruction. In addition, the bleeding effect tends to last longer in CISTA-Flow. In scenes like slider\_depth and desk\_fast, the estimated flow around the black area is close to 0, resulting in a longer smearing effect. Similar issues are observed in hand\_only, where inaccurate optical flow for the hand area affects reconstruction.


\vspace{-0.5em}
\subsubsection{Fast motion and low light condition}

\begin{figure}[tb]
     \centering
    \includegraphics[width=0.9\linewidth]{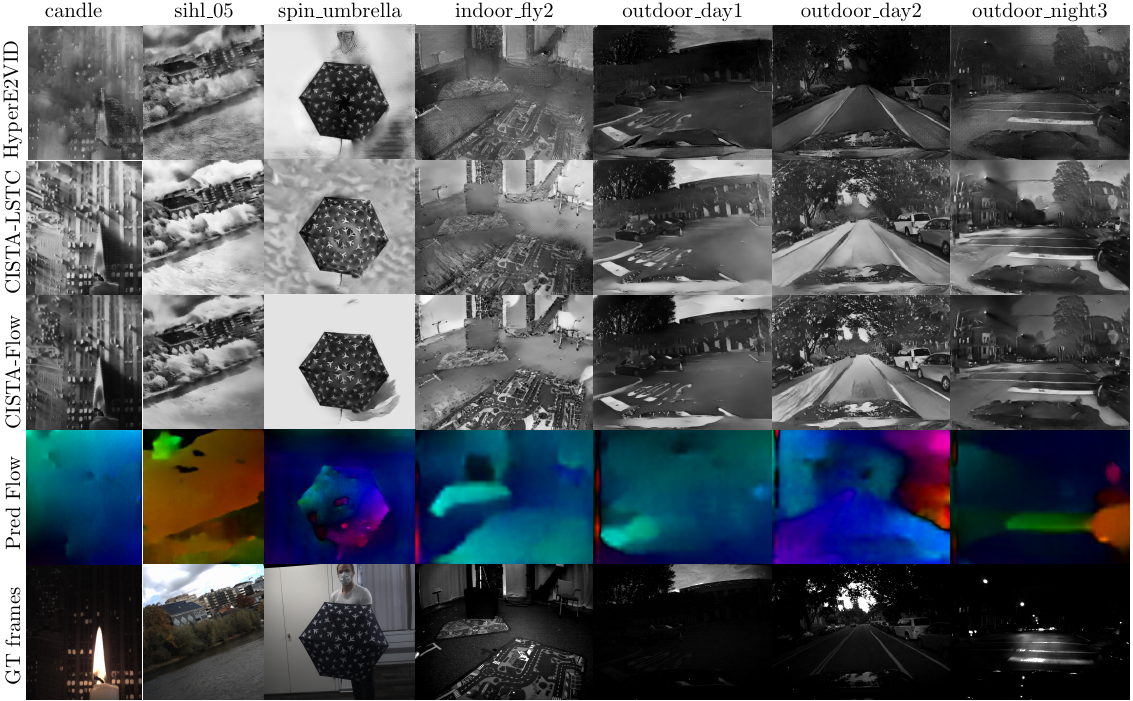}
     \caption{Reconstruction results of CISTA-LSTC and CISTA-Flow networks for challenging sequences, including scenarios with fast motion or low light condition.}
     \label{fig:compare-challenge}
     \vspace{-1em}
\end{figure}

To assess the robustness of our method, we tested CISTA-Flow on datasets with high resolution or challenging scenarios, such as scenes with fast motion or low light conditions. The reconstructed images are presented in \cref{fig:compare-challenge}. The comparison between CISTA-LSTC and CISTA-Flow images reveals motion compensation in the latter, despite some inaccuracies and inconsistencies in flow prediction. 

In high-resolution sequences from the HS-ERGB datasets, scenes featuring fast movements or rotations, like candle, sihl\_05, and spin\_umbrella, CISTA-Flow benefits from reduced motion blur and enhanced details such as umbrella patterns. HyperE2VID effectively eliminates smearing effects and produces sharper edges, as observed in spin\_umbrella, but struggles to capture certain fine details in fast-motion scenes like candle. 

For MVSEC datasets, reference images are underexposed and lack details, as presented in the last four columns in \cref{fig:compare-challenge}. HyperE2VID renders smooth frames with a dynamic range close to the reference images. However, the details are less clear compared to those produced by CISTA-Flow. In indoor\_fly2 and outdoor\_day1, edges of map and text generated by CISTA-Flow appear sharper and clearer. In low-light conditions such as outdoor\_day2 and outdoor\_night3, while HyperE2VID yields sharper edges, CISTA-Flow reveals details in dark areas. However, it exhibits more smearing effects (see outdoor\_day2) than the other two networks, due to its relatively low reconstruction quality and the resulting inaccurate flow estimation.

\subsection{Ablation Study of Different Warped Inputs}
\label{sec:results-diff-warped-inputs}
According to the analysis in \cref{sec:prob-cista}, we assume warping the previously reconstructed frame $\hat{\bm I}_{t-1}$ and sparse codes $\hat{\bm Z}^K_{t-1}$ can enhance reconstruction quality. To validate this, we compare the effects of warping different inputs. First, we train CISTA-LSTC with different inputs warped using GT flow, and then further obtain the CISTA-Flow networks with predicted flow based on these CISTA (GT Flow) networks. As shown in \cref{tab:various-warp-cista},  "No warp" is the standard CISTA-LSTC without warping. "Warp I" applies warping to the input frame, "warp I+Z" to both the input frame and sparse codes, and "warp all" extends warping to all inputs, including recurrent states $\bm a_{t-1}$ and $\bm c_{t-1}$. 

From \cref{tab:various-warp-cista}, it is evident that models with warped inputs outperform the plain CISTA-LSTC, with "warp I+Z" showing the best performance. For CISTA (GT Flow), warping only the intensity frame offers marginal improvements in reconstruction quality compared to "warp I+Z". This demonstrates the importance of warping sparse codes to enhance temporal consistency. However, for CISTA-Flow, the efficacy of warping I and Z is notably diminished due to the reduced quality of estimated flow, nearly reaching the level of "warp I". Additionally, further warping of the states is not beneficial and may even degrade reconstruction quality, as observed in "warp all" in \cref{tab:various-warp-cista}. Since the states preserve the memory from previous reconstructions, and the gates regulate the forgetting rate, warping them is unnecessary. For the flow estimation, CISTA-Flow with "warp I+Z" performs the best, exhibiting lower EPE and fewer outliers.

\begin{table}[tbp]
  \centering
  \scriptsize
  \caption{Results for CISTA (GT Flow) and corresponding CISTA-Flow networks with different warped inputs on SIM dataset. Key: \textbf{Best}}

\begin{tabular}{l|ccc|ccc|cc}
\hline
         & \multicolumn{3}{c|}{CISTA (GT Flow)} & \multicolumn{5}{c}{CISTA-Flow} \\
\hline
         & MSE$\downarrow$ & SSIM$\uparrow$ & LPIPS$\downarrow$ & MSE$\downarrow$ & SSIM$\uparrow$ & LPIPS$\downarrow$ & EPE$\downarrow$ & Out\%$\downarrow$ \\
\hline
no warp  & 0.024 & 0.650    & 0.252    &          & -        & -        & -        & - \\
\hline
warp I   & 0.028    & 0.658    & 0.241    & 0.029    & 0.652    & 0.248    & 0.672    & 3.13 \\
warp I+Z & 0.025    & \textbf{0.697} & \textbf{0.192} & \textbf{0.028} & \textbf{0.654} & \textbf{0.244} & \textbf{0.668} & \textbf{3.07} \\
warp all & \textbf{0.023}    & 0.676    & 0.205    & 0.031    & 0.620    & 0.284    & 0.691    & 3.18 \\
\hline
\end{tabular}%
  \label{tab:various-warp-cista}%
\end{table}%

\subsection{Incorporating Different Flow Networks}
\label{sec:results-flownet}
The CISTA-Flow framework supports integration with different event-based flow networks. To assess the efficacy of alternative flow networks, we incorporated ERAFT \cite{gehrig_e-raft_2021} into CISTA-Flow. ERAFT generates dense flow using just a pair of event voxel grids. Based on the pretrained ERAFT, we fine-tuned it on our dataset for 50 epochs before integrating it with CISTA-Flow. We then trained the CISTA-LSTC with the ERAFT-generated flow for 25 epochs. Comparison results between ERAFT and DCEIFlow and their corresponding CISTA-Flow networks are presented in \cref{tab:results-flownet-eraft}.

\begin{table}[tbp]
  \centering
  \scriptsize
  \caption{Comparison results of different flow networks in CISTA-Flow, where ERAFT and DCEIFlow are used. "Flow only" denotes the flow network trained independently from reconstruction. Key: \textbf{Best} results for comparison between architectures.}
  
\begin{tabular}{c|c|l|ccc|cc}
\hline
         &          &          & \multicolumn{3}{c|}{Rec}       & \multicolumn{2}{c}{Flow} \\
\cline{3-8}         &          & Dataset  & MSE$\downarrow$ & SSIM$\uparrow$ & LPIPS$\downarrow$ & EPE$\downarrow$ & Out\%$\downarrow$ \\
\hline
\multirow{4}[4]{*}{ERAFT} & Flow Only (E) & SIM      & -        & -        & -        & 0.734    & 3.66  \\
\cline{2-8}         & \multirow{3}[2]{*}{CISTA-Flow} & ECD      & \textbf{0.042 } & \textbf{0.591 } & \textbf{0.219 } & -        & - \\
         &          & HQF      & 0.041    & 0.569    & 0.258    & -        & - \\
         &          & SIM      & \textbf{0.025 } & \textbf{0.657 } & \textbf{0.243 } & 0.700    & 3.20  \\
\hline
\multirow{4}[4]{*}{DCEIFlow} & Flow Only (E+GT I) & SIM      & -        & -        & -        & \textbf{0.671 } & \textbf{2.71 } \\
\cline{2-8}         & \multirow{3}[2]{*}{CISTA-Flow} & ECD      & 0.050    & 0.586    & 0.225    & -        & - \\
         &          & HQF      & \textbf{0.030 } & \textbf{0.590 } & \textbf{0.257 } & -        & - \\
         &          & SIM      & 0.028    & 0.654    & 0.244    & \textbf{0.668 } & \textbf{3.07 } \\
\hline
\end{tabular}%
  \label{tab:results-flownet-eraft}%
\end{table}%

\begin{figure}[tbp]
    \centering
    \subfloat[DCEIFlow]{
    \begin{minipage}[c][\height][c]{0.18\linewidth}
    \centering
     \includegraphics[width=\linewidth]{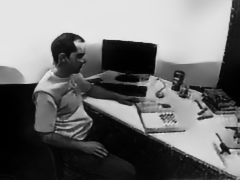}
    \end{minipage}
    }
    \subfloat[ERAFT]
    {
    \begin{minipage}[c][\height][c]{0.18\linewidth}
    \centering
     \includegraphics[width=\linewidth]{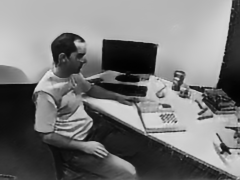}
    \end{minipage}
    }
    
    \caption{Reconstruction results of CISTA-Flow using (a) DCEIFlow and (b) ERAFT network respectively. The model with ERAFT can produce frames with reduced smearing effects. }
    \label{fig:compare-eraft}
\end{figure}

For the flow network itself, without the assistance of intensity images, ERAFT shows larger EPE and more outliers than DCEIFlow, which is consistent with findings in \cite{wan_learning_2022}. In terms of CISTA-Flow performance, DCEIFlow-generated flow remains more accurate for simulated datasets compared to ERAFT. Regarding reconstruction quality, CISTA-Flow utilizing ERAFT achieves comparable results and even slight improvements on ECD and SIM datasets. This is due to the independence of ERAFT from requiring reconstructed intensity frames as inputs. This independence ensures that inaccuracies in certain frames do not affect flow estimation and subsequent reconstructions. As shown in \cref{fig:compare-eraft}, the model with ERAFT exhibits reduced smearing effects. This experiment demonstrates the adaptability of CISTA-Flow to different flow networks and that different flow estimations could be used with CISTA-Flow in different contexts. 

\section{Conclusion}
\label{sec:conclusion}


In this study, we introduced a model-based event reconstruction network that incorporates motion compensation. The original CISTA-LSTC model, based on the assumption of shared sparse representation between input and output signals, is effective but overlooked motion displacement. To address this, we construct a CISTA-Flow network by incorporating a flow network, specifically DCEIFlow. In this system, the reconstructed frame and events are used for flow estimation, and then the estimated flow is used to warp intensity frame and sparse codes for subsequent reconstruction for better temporal consistency. An iterative training framework is also introduced for the integrated system.

Results show that our CISTA-Flow network achieves state-of-the-art reconstruction quality over other advanced networks, and produces reliable dense flow. With the assistance of motion compensation, the reconstructed frames exhibit sharper edges and finer details. However, its limitation lies in the interdependent quality of reconstruction and flow estimation, leading to extended impacts from inaccurate predictions. This issue can be alleviated by integrating event-only flow networks like ERAFT. The adaptability of CISTA-Flow to different flow networks demonstrate its potential for further improvements.

\clearpage  

%
%
\bibliographystyle{splncs04}
\bibliography{references}

\end{document}


\title{Supplementary Materials for Enhanced Event-Based Video Reconstruction with Motion Compensation} 


\author{Siying Liu\inst{1}\orcidlink{0000-0003-3533-8768} \and
Pier Luigi Dragotti \inst{1}\orcidlink{0000-0002-6073-2807}}

\authorrunning{S.~Liu and P.L.~Dragotti}

\institute{Communication and Signal Processing Group, Department of Electrical and Electronic Engineering, Imperial College London, SW7 2AZ London, UK \email{siying.liu20@imperial.ac.uk}} %

\maketitle

\section{Details of Synthetic Data}

For the synthetic dataset, we utilize the v2e simulator \cite{hu_v2e_2021} to generate events, intensity frames, and corresponding flow. Using images from MS-COCO datasets \cite{lin_microsoft_2014}, we simulate videos with arbitrary motions and 2D affine transform at adaptive frame rate. A third of the sequences only depict the background, and the others contain from 1 to 10 foreground objects moving randomly over the background. The frame rate adjusts dynamically based on motion, ensuring that the shift between two frames is no more than one pixel.

In our simulation, contrast threshold follows a Gaussian distribution $C\sim \mathcal{N}(\mu,\sigma)$, where $\mu=[0.2,0.4,0.6,0.8,1]$ randomly and $\sigma=0.03$. The values for positive and negative events are also different, where $C_n = \lambda C_p, \lambda \sim \mathcal{N}(1, 0.1)$. Lowpass filtering is executed on half of sequences with a cutoff frequency of 200 Hz, while the other half undergo filtering with a cutoff frequency drawn from $150 \times \mathcal{N}(1,0.2)$ Hz. The refractory period is set to 1ms. The leak rate and shot noise rate are 0.1 Hz and 1 Hz respectively. 


\section{Analysis of Intermediate Results}

As the CISTA-Flow is trained at different stages, we present the intermediate results in \cref{tab:interm-results}. CISTA-LSTC, the original reconstruction model, serves as a reference for comparison. CISTA (GTFlow) is the model trained with ground truth flow. For flow estimation, DCEIFlow is initially trained with events and GT frames as inputs for 50 epochs, denoted as  DCEIFlow (GT I). Subsequently, the input GT frames are replaced by reconstructed frames for an additional 20 epochs of training. This model is denoted as DCEIFlow (Rec I), with the reconstructed frames generated by CISTA (GTFlow). Finally, after iterative training based on CISTA (GTFlow) and DCEIFlow (Rec I), CISTA-Flow network with events only is obtained. 

Comparing the results of CISTA-LSTC and CISTA (GTFlow), it is evident that image quality is significantly enhanced with GT flow. The warped input image increases structural similarity between input and output, while warped sparse codes enhance temporal consistency between consecutive reconstruction. For flow estimation, using reconstructed frames instead of GT frames as inputs, we can observe a slight increase in the number of outliers from DCEIFlow (GT I) to DCEIFlow (Rec I), while the EPE is further reduced. With reconstructed frames generated by GT flow, there is only a slight performance decline, indicating increased adaptability of the flow network to sequentially reconstructed frames.



For CISTA-Flow, in the absence of ground truth flow, we observe a degradation in reconstruction performance compared to CISTA (GTFlow). In fact, on the simulated dataset, it is only marginally better than the original CISTA-LSTC in terms of SSIM and LPIPS. However, there is a significant improvement in reconstruction quality on the real datasets, as shown in Fig.~4 of the paper. In terms of flow estimation, both the EPE and the number of outliers exhibit an increase compared with DCEIFlow (Rec I).  This outcome is expected due to the lower quality of reconstructed frames and the convergence time required for each sequence, which makes reducing large errors more challenging. 


\begin{table}[htb]
  \centering
  \scriptsize
  \setlength{\tabcolsep}{1mm}
  \caption{Intermediate results on simulated dataset. Key: \textbf{Best}.}
\begin{tabular}{l|ccc|cc}
\hline
         & \multicolumn{3}{c|}{Rec}       & \multicolumn{2}{c}{Flow} \\
\hline
         & MSE$\downarrow$ & SSIM$\uparrow$ & LPIPS$\downarrow$ & \multicolumn{1}{l}{EPE$\downarrow$} & Out\%$\downarrow$ \\
\hline
CISTA-LSTC & \textbf{0.024} & 0.650    & 0.252    & -        & - \\
\hline
CISTA (GT Flow) & 0.025    & \textbf{0.697} & \textbf{0.192} & -        & - \\
DCEIFLow (GT I) & -        & -        & -        & 0.671    & \textbf{2.71} \\
DCEIFLow (Rec I) & -        & -        & -        & \textbf{0.664} & 2.83 \\
CISTA-Flow & 0.028    & 0.654    & 0.244    & 0.668    & 3.07 \\
\hline
\end{tabular}%
  \label{tab:interm-results}%
\end{table}%

\section{Further Analysis of Flow Estimation}

Ground truth flow is not available in real ECD and HQF datasets for reconstruction. We therefore use a forward warping loss (FWL) \cite{storegen_reducing_2020} to evaluate the performance of flow estimation. The FWL is computed based on the warped event image using estimated flow. Specifically, the events are warped by the optical flow at each pixel $F=(u(x, y), v(x, y))^T$ to a reference time $t^\prime$:
\begin{equation}
     I(E; F) = \left (\begin{matrix}
        x_i^\prime \\ y_i^\prime
    \end{matrix} \right ) = \left (\begin{matrix}
        x_i \\ y_i
    \end{matrix} \right ) + (t^\prime - t_i) \left (\begin{matrix}
        u(x_i, y_i) \\ v(x_i, y_i).
    \end{matrix} \right )
\end{equation}
A correct flow results in a sharper image as events are compensated. The image variance $\sigma^2(I)$ can be used to evaluate the sharpness, with a higher value suggesting better flow estimation. It is further normalized by the variance of the unwarped event image $I(E; 0)$ to obtain the FWL:
\begin{equation}
    FWL = \frac{\sigma^2(I(E;F))}{\sigma^2(I(E;0))}.
\end{equation}
When $FWL>1$, the warped event image are sharper than the unwarped image, indicating the predicted flow is better than zero flow.

\begin{figure}[tb]
     \centering
    \includegraphics[width=1.0\linewidth]{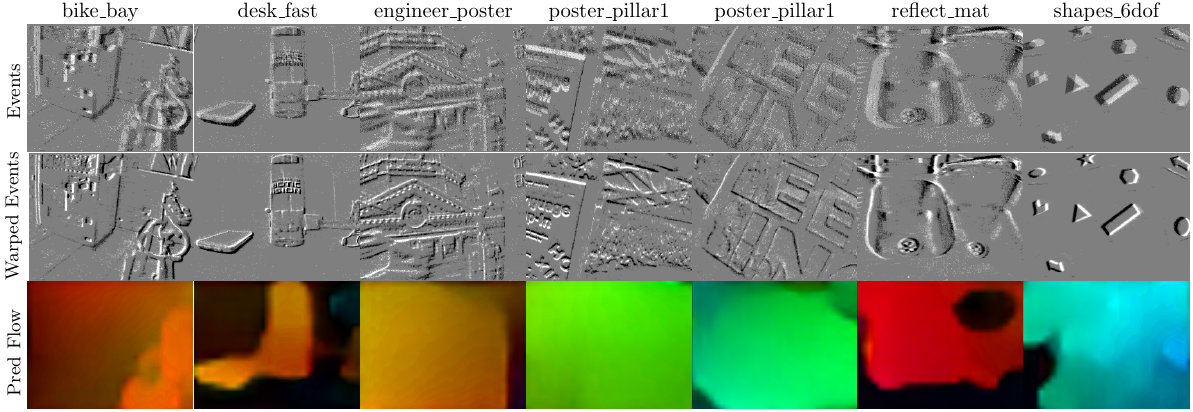}
     \caption{Comparison between the original event image and the event image warped using flow estimated by CISTA-Flow (DCEIFlow). White pixels denote positive events, while black pixels indicate negative events.} 
     \label{fig:warped_evs}
\end{figure}

\cref{tab:warped_evs} presents the average FWL scores for CISTA-Flow using various flow networks on the ECD and HQF datasets. The performance of models utilizing DCEIFlow and ERAFT are similar. In every scenario, FWL values are larget than 1, suggesting that the estimated flow enhances the sharpness of warped event images compared to those that are not warped. \cref{fig:warped_evs} illustrates a comparison between the original event images and the event images warped using flow estimated by CISTA-Flow (DCEIFlow), alongside the associated flow maps. It is obvious that the motion blur is compensated by the predicted flow, resulting in sharper edges, even though the dense flow lacks accuracy in some areas.

\begin{table}[tbp]
  \centering
  \caption{The average FWL scores of CISTA-Flow with different flow networks on ECD and HQF datasets.}
\begin{tabular}{l|cc}
\hline
FWL$\uparrow$ & ECD      & HQF \\
\hline
CISTA-Flow(DCEIFlow) & 1.29     & 1.73 \\
\hline
CISTA-Flow(ERAFT) & 1.28     & 1.74 \\
\hline
\end{tabular}%
  \label{tab:warped_evs}%
\end{table}%

\bibliography{references}
\bibliographystyle{splncs04}